# SkeletonNet: Shape Pixel to Skeleton Pixel

Sabari Nathan
Couger Inc.
sabari@couger.co.jp

Priya Kansal
Couger Inc.
priya@couger.co.jp

## Abstract

*Deep Learning for Geometric Shape Understating has organized a challenge for extracting different kinds of skeletons from the images of different objects. This competition is organized in association with CVPR 2019. There are three different tracks of this competition. The present manuscript describes the method used to train the model for the dataset provided in the first track. The first track aims to extract skeleton pixels from the shape pixels of 89 different objects. For the purpose of extracting the skeleton, a U-net model which is comprised of an encoder-decoder structure has been used. In our proposed architecture, unlike the plain decoder in the traditional U net, we have designed the decoder in the format of HED architecture, wherein we have introduced 4 side layers and fused them to one dilation convolutional layer to connect the broken links of the skeleton. Our proposed architecture achieved the F1 score of 0.77 on test data.*

## 1. Introduction

The extracted skeletons from the images are widely used in various areas like computer vision and image processing for optical character recognition [17], fingerprint recognition [28], motion detection [14], object tracking [13], etc. Skeletons are also widely used in life sciences for plant morphology [4]. Deep Learning for Geometric Shape Understanding at CVPR 2019 has organized SkelNetOn challenge. In this challenge, a pre-segmented image dataset with the corresponding skeleton representations in three tracks is provided [25]. The first track has posed the challenge of extracting the skeleton pixels from the given pre-segmented images [25][16][19][24]. We have approached this challenge as an edge detection problem and introduced a version of HED architecture in the decoder part of our proposed architecture. The rest of the sections of this manuscript describe the dataset, related work, methodology and results of the model used to secure 3[rd] place in the challenge.

## 2. Related Work

Skeleton extraction is a widely researched area in the last 10 years. However, the most recent works are mainly focused on the extracting skeleton from the RGB images [22][11], which involves segmentation or detection of the objects and extract the skeleton at the same time. Also, an extensive research is done either on edge detection [8][3][27][23] or segmentation [27][10] individually. These kinds of works do not suit fully to the present task. Some initial works are done on the extracting skeleton from the pre-segmented images [2][1][5] [9] which is similar to our task. However, most of these works are focused on the skeleton pruning to remove the unwanted branches rather than skeleton extraction. In the work done by [7], the authors introduced the boundary noise to avoid the uninformative branch creations. [15] used skeleton strength maps (SSM) which are calculated by the isotropic diffusion of the Euclidian distance transformation of binary images and their gradient. After calculating the SSM, they connected all the local maxima points of SSM with the shortest possible line to extract the skeletons. [6] approached the task of skeleton extraction as image generation model and used the generative adversarial network to extract the skeletons.

We have approached the present task as an edge detection problem and hence our work is more inspired by Holistically-nested Edge Detection (HED) model [26]. Similar to HED architecture, we have also fused the side layers into the final output layer. But to improve the performance of HED, instead of taking the output of convolution layers as side layers, we have introduced CS-SE layers at the end of each up-sampling layer and have considered the output of CS-SE layers as side layers. The detail of our approach is presented in section 3.1.

## 3. Dataset

The challenge is organized in two phases. In the development phase, 1219 images with their ground-truth for training and 242 images without ground-truth for validation are provided. In the final phase, a total of 266 test images are given. Participants are asked to submit their prediction for validation images in the development phase and test images in the final phase. For the purpose of training, we have split the training images into 80:20. While splitting, we have ensured to split object-wise, so that both training and validation sets would have all the objects. After splitting the dataset into train and validation, we found that the data is quite imbalanced. It ranges from 1image to 58 images across the 89 objects in the dataset. Hence, we have augmented the train set (975) images into 1296 images. For the purpose of augmentation, image and mask rotation from -45 degree to +45 degree are used.

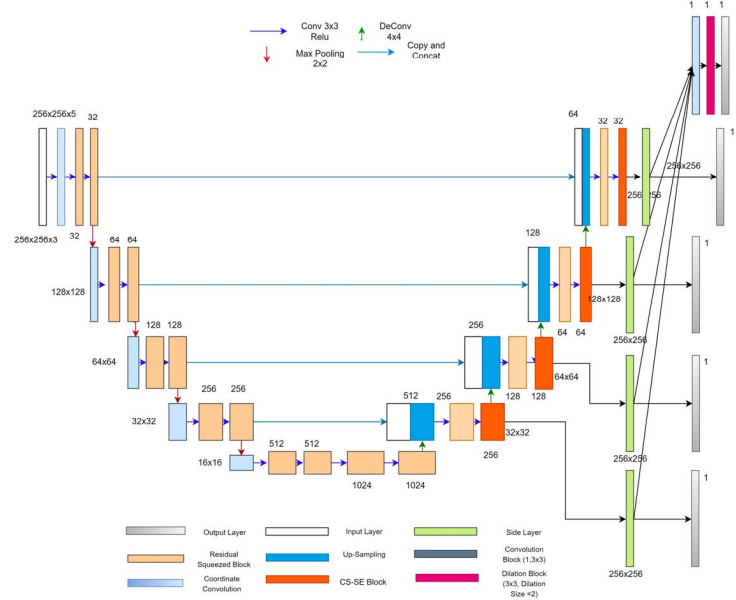

Figure 1: SkeletonNet: A detailed view of Proposed Architecture

## 4. Method:

### 4.1 Details of the Architecture:

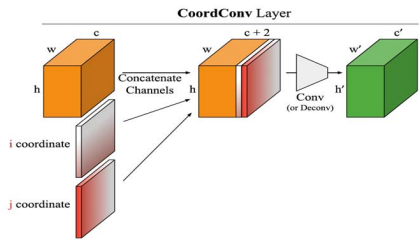

Figure 2: Coordinate convolutional layer as proposed in original paper

Unlike the plethora of classification and segmentation task, here we need to focus on the skeleton of images from the masks which is somewhat related to the problem of edge detection. In the process of extracting the skeleton from the mask of the objects, we have designed an encoder-decoder structure proposed in [20] with side layers inspired by HED architecture [26]. We have used a unique version of the HED architecture in the decoder part of U-net. The detail of the architecture is shown in Figure 1. As shown in Figure 1, before passing the image into the encoder, we have first passed the input image into a coordinate convolution layer as proposed by [18]. Coordinate convolutional layer helps the network to decide on the features related to translation invariance which further improves the generalization capacity of the model.

As suggested in the original paper, with the help of coordinate convolutional layer, spatial coordinates can be mapped with the coordinates in Cartesian space through the use of extra coordinate channels which gives the power to the model to use either complete or varying degree of translation features. Here, we have used the same two extra coordinate channels (i, j) which are suggested in the original paper. Figure 2 shows the detailed coordinate convolutional layer as given in the original paper. The coordinate channel i is a matrix in which row one is filled with all zeros, row 2 is all 1s, row three is all 2s and so on. Channel j is also similar to channel i but in this channel, the columns are filled with the numbers. Also, since we added two more channels, we have used a special residual squeezed block to extract the feature map in the encoder part (Figure 3). In our residual squeezed block, we have included the squeezed and excitation block to pass the output of the convolution layer and then have added this to the identity layer as the normal procedure of the residual block. The purpose of passing the output of residuals in the squeezed and excitation block is to prevent the overfitting caused by the extra feature maps. The squeezed and excitation block have adaptively weighed to all the feature maps [12]. As far as the decoder part is concerned, we have passed the output of up-sampling layer to the residual squeezed block. The output of the residual squeezed block is further passed to the channel squeeze and spatial excitation (CS-SE) block [21]. The CS-SE block slices its input corresponding to the spatial location (x,y) where, x ∈

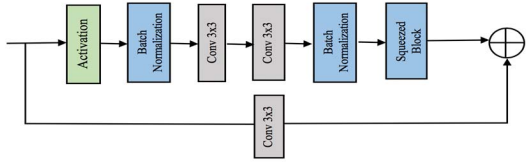

Figure 3: Residual Squeezed (RS) block used in the encoder and decoder part

As discussed before, we have used side layers inspired by the HED network. Total of 4 side layers are fused to the final output layer. The output of the fused layer is passed to the dilation layer to get the strongest features without losing the received resolution of the output of the fused layer. Further, side layers' output and the output of the final layer are then passed through a sigmoid layer individually under the supervision of ground truth (Figure 4). This approach helped us to connect the broken links of the skeleton predicted.

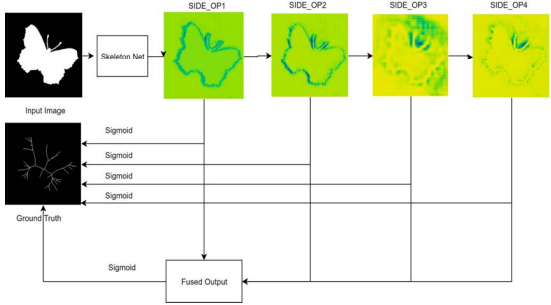

Figure 4: Side – Layers and Fused layer guidance with the Ground-truth images

### 4.2 Image Preprocessing

Images are divided by 255 to normalize the value of each pixel between 0 and 1.

### 4.3 Training

We have trained the network for five outputs which include the four side layers and one fused output layer for the skeletons of the input images. Adam optimizer is used to update the weights while training. The learning rate is initialized with 0.001 and reduced after 10 epochs to 10% if validation loss does not improve. The batch size is set to 4. The total epochs are set to 500. However, training is stopped early when the network started overfitting. The dataset is trained using Nvidia 1080 GTX GPU.

{1,2, ….H} and y ∈ {1,2, ….W}. This spatial mapping has helped the network in concentrating the meaningful features over the weak features.

### 4.4 Loss Function:

We have proposed a novel yet simple loss function. Our loss function is the sum of binary cross-entropy and Dice Loss as defined in equation 1. The network is trained to minimize the Binary loss with sigmoid activation function.

$$Loss = L + DiceLoss \quad \ldots (1)$$

Dice Loss is defined in equation (2) and L is cross-entropy loss defined in equation (3)

$$Dice\ Loss = 1 - \frac{2\sum_{i=0}^{k} y_i p_i + \varepsilon}{\sum_{i=0}^{k} y_i + \sum_{i=0}^{k} p_i + \varepsilon} \quad \ldots (2)$$

$$L = -\sum_{i=0}^{k}[y_i \log p_i + (1 - y_i)\log(1 - p_i)] \ldots (3)$$

where, $y_i$ and $p_i$ are the ground truth and the predicted skeleton images respectively. The coefficient ε is used to ensure the loss function stability by avoiding the zero value in the denominator of dice loss.

### 5. Results:

Table 1: Results of the side layer, fused layer and the ensemble output for the validation (split) data

| Output | F1-score |
|---|---|
| Side Layer 1 | 0.7708 |
| Side Layer 2 | 0.7245 |
| Side Layer 3 | 0.5832 |
| Side Layer 4 | 0.3759 |
| Fused Output | 0.7686 |
| **Ensembled** | **0.7877** |

The official metric for evaluation was F1-score. We have used the same to evaluate the results. Since the network is trained for 5 outputs, we have evaluated the output of each layer to get the best results. Table 1 shows the F1-score of all five layers. From Table 1, it is very clear that the output of first side layer is most important in the fused output. Hence, we tried to ensemble the results of the first side output layer and the fused layer. The weighted average ensemble method is used to ensemble the results.

The resulted images of this ensemble are used for final submission. The results on all the datasets are presented in Table 2.

Table 2: Results of the Proposed network: Skeleton. These results are the results of the final ensembled layers

|  | No. of Images | F1-score (ours) | F1-score (baseline [6]) |
|---|---|---|---|
| Train | 1296 | 0.8406 | - |
| Validation (Split) | 244 | 0.7877 | - |
| Validation (Original) | 242 | 0.7480 | 0.6244 |
| Test | 266 | 0.7711 | - |

Figure 5 shows the resulted images of all the outputs layers and ensembled image as well.

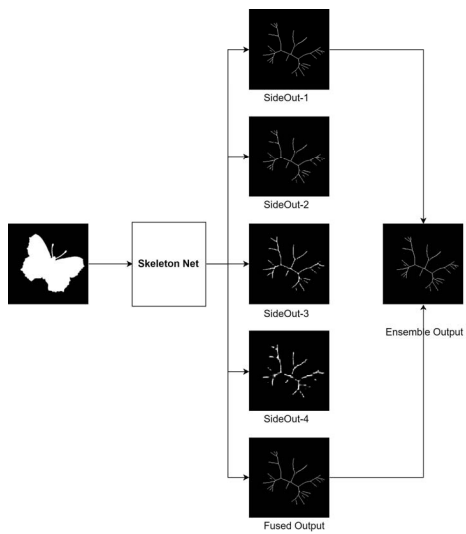

Figure 5: Side – Layers, Fused layer output and Ensembled output

## 6. Discussion:

The proposed architecture is a combination of many proven state-of-art algorithms. As discussed in section 4.1, we have used the coordinate convolutional layer to choose upon the translation features, this has helped our model to focus on more important features during the training. When compared to the plain encoder, the use of coordinate convolutional layer helped to improve the F1-score by more than 3%. However, this impact may be considered as insignificant in alone but when combined with our custom loss, it has shown the significant improvement in the learning process of the model as the F1-score have increased to 0.7686 from 0.6546 (in case of Binary Cross Entropy) on the validation (split) data.

Further, we have introduced the HED architecture i.e. the side layers in the decoder part along with the dilation layer after fusing. This has boosted up the model performance by more than 10%. Table 4 shows the F1 scores with and without side-layers in the decoder part (Table 4).

Table 4: F1 score on the validation (split) data with and without side-layer

|  | Vanilla Decoder | Decoder with Side-Layers |
|---|---|---|
| F1-score | 0.6973 | 0.7686 |

Some images from the training data along with the predicted output and ground truth are presented in Figure 6.

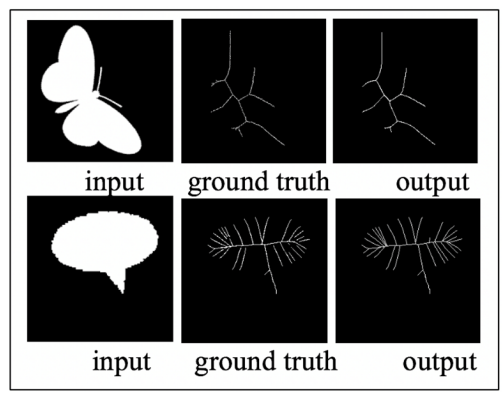

Figure 6: Illustrations of the predicted results. Results are directly compared with the ground truth images

## 7. Conclusion and Future Work:

In the present task, we have experimented a unique version of HED architecture along with the U-net structure to extract the skeleton from the pre-segmented images. Also, we have proposed a new loss function for converging the network for the best results. The present work also proves the role of side- layers in achieving the best output. As future work, we would like to explore the role of side layers in segmenting and extracting the skeleton from RGB images.

**Acknowledgement**
This work is supported by Couger Inc., Shibuya, Japan

Table 3: Impact of Using Coordination Convolutional Layer on the F1-score

|  | No Coordinate Conv Layer | | With Coordinate Conv Layer | |
|---|---|---|---|---|
| Loss Function | Binary Cross Entropy | Our Loss Function | Binary Cross Entropy | Our Loss Function |
| F1-score | 0.6546 | 0.7212 | 0.7043 | **0.7686** |